\documentclass{article} 
\pdfoutput=1
\usepackage{iclr2019_conference,times}

\usepackage[pagebackref]{hyperref}

\usepackage{amsfonts}
\usepackage{amssymb}
\usepackage{amsmath}
\usepackage{multirow}
\usepackage{makecell}
\usepackage{comment}
\usepackage{marginnote}
\usepackage{geometry}
\usepackage{physics}
\usepackage{wrapfig}
\usepackage[protrusion=true,tracking=true,kerning=true,expansion,final]{microtype}      

\usepackage[colorinlistoftodos,prependcaption,textsize=tiny]{todonotes}
\usepackage{xcolor}
\usepackage{wrapfig}

\newcommand\link[1]{\textcolor{blue}{#1}}

\usepackage[colorinlistoftodos]{todonotes}
\usepackage{graphicx}
\usepackage{subfigure}
\usepackage{booktabs} 

\usepackage{algorithm}
\usepackage{algpseudocode}

\usepackage{xcolor}

\definecolor{color_lq}{RGB}{60,180,75}
\definecolor{color_joint}{RGB}{255,225,25}
\definecolor{color_faq}{RGB}{0,128,128}
\definecolor{color_qsm}{RGB}{245, 130, 48}
\definecolor{color_dist}{RGB}{145, 30, 180}
\definecolor{color_comp}{RGB}{128, 128, 0}
\definecolor{color_pact}{RGB}{0, 0, 128}
\definecolor{color_mlq}{RGB}{175, 50, 125} 
\definecolor{color_ours}{RGB}{255,0,0}

\renewcommand{\cite}[1]{\citep{#1}}
\definecolor{mydarkblue}{rgb}{0,0.08,0.45}
\hypersetup{ %
	pdftitle={},
	pdfauthor={},
	pdfsubject={},
	pdfkeywords={},
	pdfborder=0 0 0,
	pdfpagemode=UseNone,
	colorlinks=true,
	linkcolor=mydarkblue,
	citecolor=mydarkblue,
	filecolor=mydarkblue,
	urlcolor=mydarkblue,
	pdfview=FitH}

\DeclareMathOperator{\clamp}{Clamp}
\DeclareMathOperator{\round}{Round}

\title{NICE: Noise Injection and Clamping Estimation for Neural Network Quantization}

\author{Chaim Baskin \thanks{equal contributors}, Evgenii Zheltonozhkii, Avi Mendelson \& Alexander M.Bronstein \\
	Department of Computer Science\\
	Technion\\
	Haifa, Israel\\
	\texttt{\{chaimbaskin,bron,avi.mendelson\}@cs.technion.ac.il} \\
	\texttt{evgeniizh@campus.technion.ac.il} \\
	\AND
	Natan Liss\footnotemark[1]  \\
	Department of Electrical Engineering \\
	Technion  \\
	Haifa, Israel\\
	\texttt{lissnatan@campus.technion.ac.il} \\
	\AND
	Yoav Chai, Eli Schwartz \& Raja Giryes \\
	School of Electrical Engineering \\
	Tel-Aviv University\\ 
	Tel-Aviv, Israel\\
	\texttt{\{yoavchai1@mail,eliyahus@mail,raja@tauex\}.tau.ac.il}  \\
}

\newcommand{\uniqname}{NICE } 
\iclrfinalcopy 

\begin{document}

	\maketitle

	\begin{abstract}
		Convolutional Neural Networks (CNN) are very popular in many fields including computer vision, speech recognition, natural language processing, to name a few.
		Though deep learning leads to groundbreaking performance in these domains, the networks used are very demanding computationally and are far from real-time even on a GPU, which is not power efficient and therefore does not suit low power systems such as mobile devices.
		To overcome this challenge, some solutions have been proposed for quantizing the weights and activations of these networks, which accelerate the runtime significantly. Yet, this acceleration comes at the cost of a larger error. The \uniqname method proposed in this work trains quantized neural networks by noise injection and a learned clamping, which improve the accuracy.
		This leads to state-of-the-art results on various regression and classification tasks, e.g.,
		ImageNet classification with architectures such as ResNet-18/34/50 with low as 3-bit weights and activations. We implement the proposed solution on an FPGA to demonstrate its applicability for low power real-time applications.   Our code is publicly available at \link{https://github.com/Lancer555/NICE}
		
	\end{abstract}

	\section{Introduction}
	Deep neural networks have established themselves  as an important tool in the machine learning arsenal. They have shown spectacular success in a variety of tasks in a broad range of fields such computer vision, computational and medical imaging,  signal, image, speech and language processing \cite{6296526,Lai:2015:RCN:2886521.2886636,DBLP:journals/corr/ChenPK0Y16}.
	
	However, while deep learning models' performance is impressive, the computational and storage requirements of both training and inference are harsh. For example, ResNet-50 \cite{He_2016_CVPR}, a popular choice for image detection, has 98 MB parameters and requires 4 GFLOPs of computations for a single inference.  In many cases, the devices do not have such a big amount of resources, which makes deep learning infeasible in smart phones and the Internet of Things (IoT).
	
	In attempt to solve these problems, many researchers have recently came up with less demanding models, often at the expense of more complicated training procedure. Since the training is usually performed on servers with much larger resources, this is usually an acceptable trade-off.
	
	One prominent approach is to quantize the networks. The default choice for the data type of the neural networks' weights and feature maps (activations) is 32-bit (single precision) floating point. \citet{gupta2015deep} have shown that quantizing the pre-trained weights to $16$-bit fixed point have almost no effect on the accuracy of the networks. Moreover, minor modifications allow performing an integer-only 8-bit inference with reasonable performance degradation \cite{jacob2017quantization}, which is utilized in DL frameworks, such as TensorFlow. One of the current challenges in network quantization  is reducing the precision even further, up to 1-5 bits per value. In this case, straightforward techniques result in unacceptable quality degradation.

	\textbf{Contribution.} This paper introduces a novel simple approach denoted \uniqname (noise injection and clamping estimation) for neural network quantization that relies on the following two easy to implement components: (i) Noise injection during training that emulates the quantization noise introduced at inference time; and (ii) Statistics-based initialization of parameter and activation clamping, for faster model convergence. In addition, activation clamp is learned during train time. We also propose integer-only scheme for an FPGA on regression task \cite{schwartz2018deepisp}.

	
	
	
	
	
	
	Our proposed strategy for network training lead to an
	improvement over the state-of-the-art quantization techniques in the performance vs.\ complexity tradeoff. Unlike several leading methods, our approach can be applied directly to existing architectures without the need to modify them at training (as opposed, for example, to the teacher-student approaches \cite{polino2018model} that require to train a bigger network, or the XNOR networks \cite{rastegari2016xnor}
	that typically increase the number of parameters by a significant factor in order to meet accuracy goals). 
	
	Moreover, our new technique allows quantizing all the parameters in the network to fixed point (integer) values. This include the batch-norm component that is usually not quantized in other works. Thus, our proposed solution makes the integration of neural networks in dedicated hardware devices such as FPGA and ASIC easier. As a proof-of-concept, we present also a case study of such an implementation on hardware.

	\section{Related work}
	\label{gen_inst}
	
{\bf Expressiveness based methods.}	The quantization of neural network to extremely low-precision representations (up to 2 or 3 possible values) was actively studied in recent years  \cite{rastegari2016xnor,hubara2016quantized,mishra2017wrpn,ZhangYangYeECCV2018}. To overcome the accuracy reduction, some works proposed to use a wider network \cite{zhu2016trained,polino2018model,DBLP:journals/corr/abs-1805-11046}, which compensates the expressiveness reduction of the quantized networks network. For example, 32-bit feature maps were regarded as 32 binary ones. Another way to improve expressiveness, adopted by \citet{zhu2016trained} and \citet{zhou2017incremental} is to add a linear scaling layer after each of the quantized layers.
	
{\bf Keeping full-precision copy of quantized weights.}	Lately, the most common approach to training a quantized neural network \cite{hubara2016binarized,hubara2016quantized,zhou2016dorefa,rastegari2016xnor,cai2017deep} is keep two sets of weights --- forward pass is performed with quantized weights, and updates are performed on full precision ones, i.e., approximating gradients with straight-through estimator (STE) \cite{bengio2013estimating}. For quantizing the parameters, either stochastic or deterministic function can be used.
	
{\bf Distillation.}	One of the leading approaches used today for quantization relies on the idea of distillation \cite{hinton2015distilling}. In distillation a teacher-student setup is used, where the teacher is either the same or a larger full precision neural network and the student is the quantized one. The student network is trained to imitate the output of the teacher network. This strategy is successfully used to boost the performance of existing quantization methods \cite{mishra2017apprentice,polino2018model,2018arXiv180805779J}.
	
{\bf Model parametrization.}	\citet{ZhangYangYeECCV2018} proposed to represent the parameters with learned basis vectors that allow acquiring an optimized non-uniform representation. In this case MAC operations can be computed with bitwise operations. \citet{choi2018pact} proposed to learn the clamping value of the activations to find the balance between clamping and quantization errors. In this work we also learn this value but with the difference that we are learning the clamps value directly using STE back-propagation method without any regulations on the loss.    
\citet{2018arXiv180805779J} created a more complex parametrization of both weights and activations, and approximated them with symmetric piecewise linear function, learning both the domains and the parameters directly from the loss function of the network.
	
{\bf Optimization techniques.} \citet{zhou2017incremental} and \citet{dong2017learning} used the idea of not quantizing all the weights simultaneously but rather gradually increasing the number of quantized weights to improve the convergence.
\citet{mckinstry2018disc} demonstrated that 4-bit fully integer neural networks can achieve full-precision performance by applying simple techniques to combat varience of gradients: larger batches and proper learning rate annealing with longer training time. However, 8-bit and 32-bit integer representations were used for the multiplicative (i.e., batch normalization) and additive constants (biases), respectively.	
	
{\bf Generalization bounds.}	Interestingly, quantization of neural networks have been used recently as a theoretical tool to understand better the generalization of neural networks. It has been shown that while the generalization error does not scale with the number of parameters in over-parameterized networks, it does so when these networks are being quantized \cite{arora2018stronger}. 

	\section{Method}
	\label{method}

In this work we propose a training scheme for quantized neural networks designed for fast inference on hardware with integer-only arithmetic. To achieve maximum performance, we apply a combination of several well-known as well as novel techniques.
Firstly, in order to emulate the effect of quantization, we inject additive random noise into the network weights. Uniform noise distribution is known to approximate well the quantization error for fine quantizers; however, our experiments detailed in the sequel show that it is also suitable for relatively coarse quantization (Appendix \ref{sec:quant_err_dist}). Furthermore, some amount of random weight perturbation seems to have a regularization effect beneficial for the overall convergence of the training algorithm. 
Secondly, we use a gradual training scheme to minimize the perturbation of network parameters performed simultaneously. In order to give the quantized layers as much gradient updates as possible, we used the STE approach to pass the gradients to the quantized layers. After the gradual phase, the whole network is quantized and trained for a number of fine-tuning epochs. 
Thirdly, we propose to clamp both the activations and the weights in order to reduce the quantization bin size (and, thus, the quantization error) at the expense of some sacrifice of the dynamic range. The clamping values are initialized using the statistics of each layer. In order to truly optimize the tradeoff between the reduction of the quantization error vs that of the dynamic range, we learn optimal clamping values by
defining a loss on the quantization error. 

Lastly, following common approach proposed by \citet{zhou2016dorefa}, we don't quantize first and last layers of the networks, which have significantly higher impact on network performance.
%
The remainder of the section details these main ingredients of our method.

We propose to inject uniform additive noise to weights and biases during model training to emulate the effect of quantization incurred at inference.    
Prior works have investigated the behavior of quantization error \cite{sripad1977necessary, gray1990quantization} and concluded that in sufficiently fine-grain quantizers it can be approximated as a uniform random variale. We have observed the same phenomena and empirically verified it for weight quantization as coarse as $5$ bits.  

The advantage of the proposed method is that the updates performed during the backward pass immediately influence the forward pass, in contrast to strategies that directly quantize the weights, where small updates often leave them in the same bin, thus,  effectively unchanged. 
    
	In order to achieve a dropout-like effect in the noise injection, we use a Bernoulli distributed mask $M$, quantizing part of the weights and adding noise to the others. From empirical evidence, we chose $M \sim \mathrm{Ber}(0.05)$ as it gave the best results for the range of bitwidths in our experiments. Instead of using the quantized value $\hat{w} = \mathcal{Q}_{\Delta}(w)$ of a weight $w$ in the forward pass, $\hat{w} = (1-M)  \mathcal{Q}_{\Delta}(w) + M (w-e)$ is used with $e \sim \mathrm{Uni}(-\Delta/2,\Delta/2)$, where $\Delta$ denotes size of the quantization bin.

    \subsection{{Gradual quantization} }
    \label{sec:gradual}
    
    In order to improve the scalability of the method for deeper networks, it is desirable to avoid the significant change of the network behavior due to quantization. Thus, we start from gradually adding a subset of weights to the set of quantized parameters, allowing the rest of the network to adapt to the changes.
    
	The gradual quantization is performed in the following way: the network is split into $N$ equally-sized blocks of layers $\{B_1,...,B_N\}$. At the $i$-{th} stage, we inject the noise into the weights of the layers from the block $B_{i}$. The previous blocks $\{B_1,...,B_{i-1}\}$ are quantized, while the following blocks $\{B_{i+1},...,B_N\}$ remain at full precision. 
We apply the gradual process only once, i.e., when the $N$-{th} stage finishes, in the remaining training epochs we quantize and train all the layers using the STE approach.
	
	This gradual increasing of the number of quantized layers is similar to the one proposed by \citet{AAAI1816479}. This gradual process reduces, via the number of parameters, the amount of simultaneously injected noise and improves convergence. Since we start from the earlier blocks, the later ones have an opportunity adapt to the quantization error affecting their inputs and thus the network does not change drastically during any phase of quantization. After finishing the training with the noise injection into the block of layers $B_N$, we continue the training of the fully quantized network for several epochs until convergence. 
    In the case of a pre-trained network destined for quantization, we have found that the optimal block size is a single layer with the corresponding activation, while using more than one epoch of training with the noise injection per block does not improve performance. 
	
	\subsection{Clamping and quantization}
    \label{sec:param}
	
In order to quantize the network weights, we clamp their values in the range $[-c_w, c_w]$:
	\begin{eqnarray}
	w_c = \clamp(w, -c_{w}, c_{w}) = \max\qty(-c_{w}, \min\qty(x, c_{w})).
	\end{eqnarray}
   The parameter $c_w$ is defined per layer and is initialized with
$c_{w}=\text{mean}(w) + \beta \times \text{std}(w)$, where $w$ are the weighs of the layer and $\beta$ is a hyper-parameter. Given $c_{w}$, we uniformly quantize the clamped weight into $B_w$ bits according to
	$$\hat{w} = \left[w_c \frac{2^{B_w-1}-1}{c_{w} }\right] \frac{c_{w}}{2^{B_w-1}-1 },$$
    where $[\cdot]$ denotes the rounding operation. 
    
The quantization of the network activations is performed in a similar manner. The conventional ReLU activation function in CNNs is replaced by the clamped ReLU,
	\begin{eqnarray}
	a_c = \clamp(a,0,c_{a}),
	\end{eqnarray}
    where $a$ denotes the output of the linear part of the layer, $a_c$ is nonnegative value of the clamped activation prior to quantization, and $c_{a}$ is the clamping range.	
	The constant $c_{a}$ is set as a local parameter of each layer and is learned  with the other parameters of the network via backpropagation. We used the initialization
	$c_a=\text{mean}(a) + \alpha \times \text{std}(a)$ with the statistics computed on the training dataset and $\alpha$ set as a hyper-parameter.
	
A quantized version of the truncated activation is obtained by quantizing $a_c$ uniformly to $B_a$ bits,
	\begin{eqnarray}
	\hat{a} = \left[a_c \frac{2^{B_a}-1}{c_{a} }\right] \cdot \frac{c_{a}}{2^{B_a}-1}.
	\end{eqnarray}
	Since the $\round$ function is non-differentiable, we use the STE approach to propagate the gradients through it to the next layer. For the update of $c_a$, we calculate the derivative of $\hat{a}$ with respect to $c_a$ as
	\begin{eqnarray}
	\pdv{ \hat{a} }{a_c}=
	\begin{cases}
	1, & a_c \in [0,c_{a}]  \\
    0, & \mathrm{otherwise}.
	\end{cases}
	\end{eqnarray}

Additional analysis of the clamping parameter convergence is presented in Appendix \ref{act_clamp_app}.

	The quantization of the layer biases is more complex, since their scale depends on the scales of both the activations  and the weights. For each layer, we initialize the bias clamping value as
	\begin{equation}
c_b= \qty(\underbrace{\frac{c_{a}}{2^{B_a}-1 }}_\text{Activation scale} \cdot \underbrace{\frac{c_{w}}{2^{B_w-1}-1 }}_\text{Weight scale})\cdot\qty(\underbrace{{2^{B_b-1}-1}}_\text{Maximal bias value}), \label{eq:1}
	\end{equation}
where $B_b$ denotes the bias bitwidth. 
The biases are clamped and quantized in the same manner as  the weights.

	\section{Experimental results}
	\label{others}

	To demonstrate the effectiveness of our method, we implemented it in PyTorch and evaluated on image classification datasets (ImageNet and CIFAR-10) and a regression scenario (the MSR joint denoising and demosaicing dataset \cite{msrdemosaicing2014}). The CIFAR-10 results are presented in Appendix \ref{cifar_app}. In all the experiments, we use a pre-trained FP32 model, which is then quantized using \uniqname.

	\subsection{ImageNet}
	
For quantizing the ResNet-18/34/50 networks for ImageNet, we fine-tune a given pre-trained network using \uniqname.
    We train a network for a total of 120 epochs, following the gradual process described in Section~\ref{sec:gradual} with the number of stages $N$ set to the number of trainable layers. We use an SGD optimizer with learning rate is $10^{-4}$, momentum $0.9$ and weight decay $4\times 10^{-5}$. 
    
	Table \ref{tab_imagenet} compares \uniqname with other leading approaches to low-precision quantization \cite{2018arXiv180805779J, choi2018pact, ZhangYangYeECCV2018, mckinstry2018disc}. Various quantization levels of the weights and activations are presented. As a baseline, we use a pre-trained full-precision model.
    
		Our approach achieves state-of-the-art results for 4 and 5 bits quantization and comparable results for 3 bits quantization, on the different network architectures. Moreover, notice that our results for the 5,5 setup, on all the tested architectures, have slightly outperformed the FAQ 8,8 results.

       \begin{wrapfigure}[25]{R}{0.6\textwidth}
		\begin{center}
\vspace{-0.9cm}		\centerline{\includegraphics[width=\linewidth]{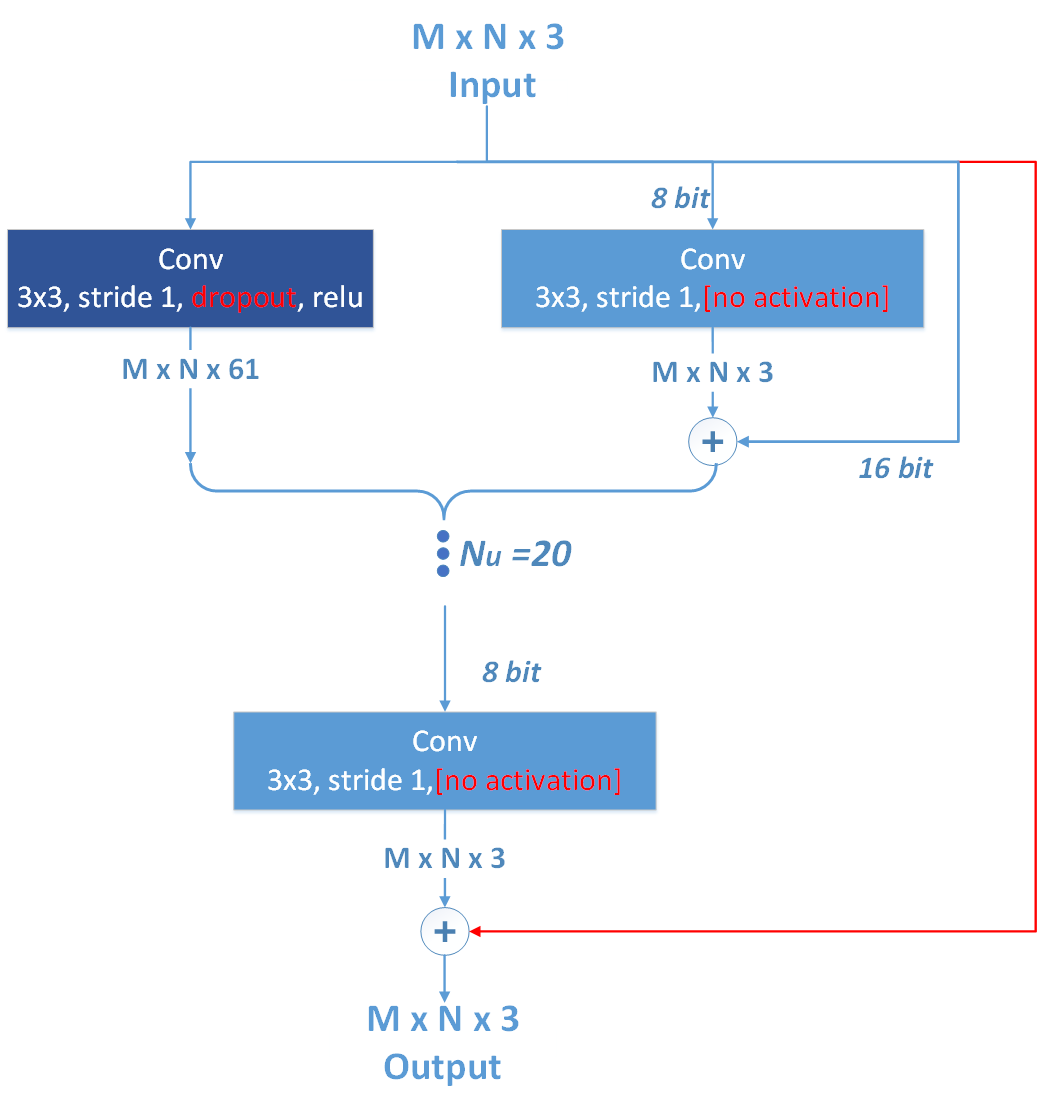}}
			\caption{
				 Model used in denoising/demosaicing experiment}
			\label{denosing_figure}
		\end{center}
		\vskip -0.2in
\end{wrapfigure}

	\subsection{Regression - joint denoising and demosaicing}
	
 	\begin{table}[ht]
		\caption{PSNR [dB] results on  joint denoising and demosaicing for different bitwidths.}
        \vspace{0.1cm}
		\label{tab_denosing_comparison}
		\begin{center}
			\begin{small}
				
				\vskip -0.1in
				\begin{tabular}{llcccc}
					\toprule
					
					&
					\bf{Bits}        &
					\bf{Bits}        &
					\bf{Bits}        &
					\bf{Bits}        &
					\bf{Bits}
					\\

					\bf{Method}      &
					\bf{(w=32,a=32)} &
					\bf{(w=4,a=8)}   &
					\bf{(w=4,a=6)}   &
					\bf{(w=4,a=5)}   &
					\bf{(w=3,a=6)}
					\\
					
					\midrule
					{\uniqname (Ours)}           & 39.696 & 39.456 & 39.332 & 39.167 & 38.973 \\
					{WRPN (our experiments)}           & 39.696 & 38.086 & 37.496 & 36.258 & 36.002 \\
					
					\bottomrule
				\end{tabular}
			\end{small}
		\end{center}
		\vskip -0.1in
	\end{table}

	In addition to the classification tasks, we apply \uniqname on a regression task, namely joint image denoising and demosaicing. The network we use is the one proposed in \cite{schwartz2018deepisp}.     
We slightly modify it by adding to it Dropout with $p=0.05$, removing the $\tanh$ activations and adding skip connections between the input and the output images. These skip connections improve the quantization results as in this case the network only needs to learn the necessary modifications to the input image.
Figure \ref{denosing_figure} shows the whole network, where the modifications are marked in red. The three channels of the input image are quantized to 16 bit, while the output of each convolution, when followed by an activation, are quantized to 8 bits (marked in Fig. \ref{denosing_figure}). The first and last layers are also quantized. 

We apply \uniqname on a full precision pre-trained network for 500 epochs with Adam optimizer with learning rate of $3 \cdot 10^{-5}$. The data is augmented with random horizontal and vertical flipping. Since we are not aware of any  other work of quantization for this task, we implemented WRPN \cite{mishra2017wrpn} as a baseline for comparison. Table~\ref{tab_denosing_comparison} 
reports the test set PSNR for the MSR dataset \cite{msrdemosaicing2014}.
It can be clearly seen that \uniqname achieves significantly better results than WRPN, especially for low weight bitwidths.

		\begin{table}[t]
		\centering
		\caption{ ImageNet comparison. We report top-1, top-5 accuracy on ImageNet compared with state-of-the-art prior methods. For each DNN architecture, rows are sorted in number of bits.Baseline results were token from PyTorch model zoo. Compared methods:  \textcolor{color_joint}{JOINT} \cite{2018arXiv180805779J}, \textcolor{color_pact}{PACT} \cite{choi2018pact}, \textcolor{color_lq}{LQ-Nets} \cite{ZhangYangYeECCV2018}, \textcolor{color_faq}{FAQ} \cite{mckinstry2018disc}}
		\begin{tabular}{ccccc}
			\toprule
			\textbf{Network} & \textbf{Method}                    & \textbf{Precision (w,a)} & \textbf{Accuracy (\% top-1)} & \textbf{Accuracy (\% top-5)} \\ \midrule
			ResNet-18        & baseline                           & 32,32                    & 69.76                        & 89.08                        \\
			ResNet-18        & \textcolor{color_faq}{FAQ}         & 8,8                      & 70.02                        & 89.32                        \\
			ResNet-18        & \textcolor{color_ours}{\uniqname(Ours)} & 5,5                      & \textbf{70.35}               & \textbf{89.8}                \\
			ResNet-18        & \textcolor{color_pact}{PACT}       & 5,5                      & 69.8                         & 89.3                         \\ \noalign{\vskip 1mm}   
			ResNet-18        & \textcolor{color_ours}{\uniqname(Ours)} & 4,4                      & 69.79                        & \textbf{89.21}               \\
			ResNet-18        & \textcolor{color_joint}{JOINT}     & 4,4                      & 69.3                         & -                            \\

ResNet-18        & \textcolor{color_pact}{PACT}       & 4,4                      & 69.2                         & 89.0                         \\
			ResNet-18        & \textcolor{color_faq}{FAQ}         & 4,4                      & \textbf{69.81 }              & 89.10                        \\
			ResNet-18        & \textcolor{color_lq}{LQ-Nets}      & 4,4                      & 69.3                         & 88.8                         \\  \noalign{\vskip 1mm} 
			ResNet-18        & \textcolor{color_joint}{JOINT}     & 3,3                      & \textbf{68.2}                & -                            \\
			ResNet-18        & \textcolor{color_ours}{\uniqname(Ours)} & 3,3                      & 67.68                         & \textbf{88.2}               \\
			ResNet-18        & \textcolor{color_lq}{LQ-Nets}      & 3,3                      & \textbf{68.2 }               & 87.9                         \\
			ResNet-18        & \textcolor{color_pact}{PACT}       & 3,3                      & 68.1                         & \textbf{88.2}                \\\midrule
			
			ResNet-34        & baseline                           & 32,32                    & 73.30                        & 91.42                        \\
			ResNet-34        & \textcolor{color_faq}{FAQ}         & 8,8                      &73.71               & \textbf{91.63}               \\
			ResNet-34        & \textcolor{color_ours}{\uniqname(Ours)} & 5,5                      & \textbf{73.72}                         & 91.60                          \\  \noalign{\vskip 1mm} 
			ResNet-34        & \textcolor{color_ours}{\uniqname(Ours)} & 4,4                      & \textbf{ 73.45 }             & \textbf{91.41}               \\
			ResNet-34        & \textcolor{color_faq}{FAQ}         & 4,4                      & 73.31                        & 91.32                        \\  \noalign{\vskip 1mm} 
			ResNet-34        & \textcolor{color_lq}{LQ-Nets}      & 3,3                      & \textbf{71.9}                         & 88.15                        \\
			ResNet-34        & \textcolor{color_ours}{\uniqname(Ours)} & 3,3                      & 71.74                         & \textbf{90.8}                         \\\midrule

			ResNet-50        & baseline                           & 32,32                    & 76.15                        & 92.87                        \\
			ResNet-50        & \textcolor{color_faq}{FAQ}         & 8,8                      & 76.52                        & 93.09                        \\
			ResNet-50        & \textcolor{color_pact}{PACT}       & 5,5                      & 76.7                         & 93.3                         \\
			ResNet-50        & \textcolor{color_ours}{\uniqname(Ours)} & 5,5                      & \textbf{76.73 }               & \textbf{93.31 }               \\  \noalign{\vskip 1mm} 
			ResNet-50        & \textcolor{color_ours}{\uniqname(Ours)} & 4,4                      & \textbf{ 76.5}               & \textbf{93.3}                \\
			ResNet-50        & \textcolor{color_lq}{LQ-Nets}      & 4,4                      & 75.1                         & 92.4                         \\
			ResNet-50        & \textcolor{color_pact}{PACT}       & 4,4                      & \textbf{76.5 }               & 93.2                         \\
			ResNet-50        & \textcolor{color_faq}{FAQ}         & 4,4                      & 76.27                        & 92.89                        \\  \noalign{\vskip 1mm} 
 			ResNet-50        & \textcolor{color_ours}{\uniqname(Ours)} & 3,3                      & 75.08                          & 92.35                          \\
			ResNet-50        & \textcolor{color_pact}{PACT}       & 3,3                      & \textbf{75.3 }               & \textbf{92.6}                \\
			ResNet-50        & \textcolor{color_lq}{LQ-Nets}      & 3,3                      & 74.2                & 91.6                \\
			
			\bottomrule
		\end{tabular}
		\label{tab_imagenet}
	\end{table}
	\subsection{Ablation study}
	In order to show the importance of each part of our \uniqname method, we use ResNet-18 on ImageNet. Table~\ref{tab_ablation} reports the accuracy for various combinations of the \uniqname components. Notice that for high bitwidths, i.e., 5,5 the noise addition and gradual training contribute to the accuracy more than the clamp learning. This happens since (i) the noise distribution is indeed uniform in this case as we show in Appendix~\ref{sec:quant_err_dist}; and (ii) the relatively high number of activation quantization levels almost negates the effect of  clamping.
	For low bitwidths, i.e 3,3, we observe the opposite. The uniform noise assumption is no longer accurate. Moreover, due to the small number of bits, clamping the range of values becomes  more significant.

	\begin{table}[ht]
		\caption{Ablation study of \uniqname scheme. Accuracy (\% top-1) for ResNet-18  on ImageNet for different setups}
		\label{tab_ablation}
		\begin{center}
			\begin{small}
				
				\begin{tabular}{cccc}
					\toprule
					
					\bf{Noise+Gradual training}       &
					\bf{Activation clamping learning} &
					\bf{Accuracy on 5,5 [W,A]}        &
					\bf{Accuracy on 3,3 [W,A]}
					\\
					
					\midrule
					-                                 & -          & 69.72 & 66.51 \\
					-                                 & \checkmark & 69.9  & 67.2 \\
					\checkmark                        & -          & 70.25 & 66.7 \\
\checkmark                        & \checkmark & 70.3  & 67.68 \\

					\bottomrule
				\end{tabular}
			\end{small}
		\end{center}
		\vskip -0.1in
	\end{table}
	
	\section{Hardware Implementation}
	\label{hardware}
	
	\subsection{optimizing quantization flow for hardware inference}
	Our quantization scheme can fits an FPGA implementation well for several reasons. Firstly, uniform quantization of both the weights and activation induces  uniform steps between each quantized bin. This means that we can avoid the use of a resource costly code-book (look-up table) with the size $B_a \times B_w \times B_a$, for each layer. This also saves calculation time.
	
	Secondly, our method enables having an integer-only arithmetic. In order to achieve that, we start, following \eqref{eq:1}, by representing each activation and network parameter in the form of $ X = N\times S$, where N is the integer code and S is a pre-calculated scale. We then reformulate the scaling factors $S$ into the form $ \hat{S} = q\times 2^{p}$, where $q \in \mathbb{N}, p \in \mathbb{Z}$. Practically, we found that it is sufficient to constrain these values to $q \in [1, 256]$ and $p \in [-32, 0]$ without an accuracy drop .This representation allows the replacement of hardware costly floating point operations by a combination of cheap shift operations and integer arithmetics.

	\subsection{Hardware Flow}
    
\begin{figure*}[t]
		\vskip 0in
		\begin{center}
			\centerline{\includegraphics[width=0.5\linewidth]{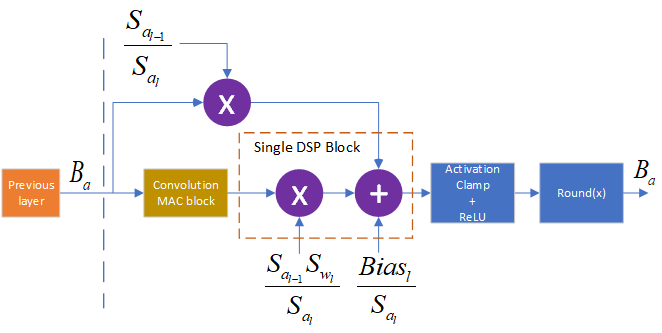}}
			\vspace{-0.3cm}
			\caption{ Residual block in hardware  }
			\label{hardware_uniq_figure}
		\end{center}
		\vskip -0.2in
	\end{figure*}

	In the hardware implementation, for both the regression and the classification tasks, we adopt the PipeCNN \cite{DBLP:journals/corr/WangAX16} implementation released by the authors.\footnote{\href{https://github.com/doonny/PipeCNN}{https://github.com/doonny/PipeCNN}}
	In this implementation, the FPGA is programmed with an image containing data moving, convolution and a pooling kernels.
	Layers are calculated sequentially.
	Figure \ref{hardware_uniq_figure} illustrates the flow of feature maps in the residual block from a previous layer to the next one. $S{a_{i}}, S{w_{i}}$ are the activations and weights scale factors of layer $i$, respectively. All these factors are calculated off-line and are loaded to the memory along with the rest of the parameters. Note that we use the FPGA for inference only.
	
    We have compiled the OpenCL kernel to Intel's Arria 10
    FPGA and run it with the regression architecture in Fig~\ref{denosing_figure}. Weights are quantized to 4 bits, activations to 8 bits, and biases and the input image to 16 bits. The resource utilization is amounted to  222K LUTs, 650 DSP Blocks and 35.3 Mb of on-chip RAM. With the maximum clock frequency of 240MHz, the processing of a single image takes 250ms. In terms of power, the FPGA requires $30W$ while an NVIDIA Titan X GPU requires $160W$.
    From standard hardware design practices, we can project that a dedicated ASIC manufactured using a similar process would be much more efficient by at least one order of magnitude.    
	

				
					
					
					

	\section{Conclusion}

	We introduced \uniqname -- a training scheme for quantized neural networks. The scheme is based on using uniform quantized parameters, additive uniform noise injection and learning the quantization clamping range. The scheme is amenable to efficient training by back propagation in full precision arithmetic.
One advantage of \uniqname is the ease of its implementation on existing networks. In particular, it does not requires changes in the architecture of the network, such as increasing the number of filters as required by some previous works. Moreover, \uniqname can be used for various types of tasks such as classification and regression.

	We report state-of-the-art results on ImageNet for a range of bitwidths and network architectures. Our solution outperforms current works on both the 4,4 and 5,5 setups, for all tested architectures, including non-uniform solutions such as \cite{ZhangYangYeECCV2018}. It shows comparable results in the 3,3 setup.
	
    We have shown that quantization error for 4 and 5 bits distributes uniformly, which explains the larger success of our method in  these bitwidths compared to the case of 3 bits. This implies that the results for less than 4 bits may be further improved by adding  non-uniform noise to the parameters. However, the 4 bit quantization  is of special interest since, being a power of 2, it is considered more hardware friendly, and INT4 matrix multiplications are supported by Tensor Cores in recently announced inference-oriented Nvidia's Tesla GPUs.

	
	\subsubsection*{Acknowledgments}
	The research was funded by ERC StG RAPID and ERC StG SPADE. 
    
	 \clearpage 
	\bibliography{iclr2019_conference}

\begin{thebibliography}{38}
\providecommand{\natexlab}[1]{#1}
\providecommand{\url}[1]{\texttt{#1}}
\expandafter\ifx\csname urlstyle\endcsname\relax
  \providecommand{\doi}[1]{doi: #1}\else
  \providecommand{\doi}{doi: \begingroup \urlstyle{rm}\Url}\fi

\bibitem[Arora et~al.(2018)Arora, Ge, Neyshabur, and Zhang]{arora2018stronger}
Sanjeev Arora, Rong Ge, Behnam Neyshabur, and Yi~Zhang.
\newblock Stronger generalization bounds for deep nets via a compression
  approach.
\newblock In \emph{international conference on machine learning (ICML)}, 2018.

\bibitem[Aydonat et~al.(2017)Aydonat, O'Connell, Capalija, Ling, and
  Chiu]{DBLP:journals/corr/AydonatOCLC17}
Utku Aydonat, Shane O'Connell, Davor Capalija, Andrew~C. Ling, and Gordon~R.
  Chiu.
\newblock An opencl\texttrademark deep learning accelerator on arria 10.
\newblock In \emph{Proceedings of the 2017 ACM/SIGDA International Symposium on
  Field-Programmable Gate Arrays}, FPGA '17, pp.\  55--64, New York, NY, USA,
  2017. ACM.
\newblock ISBN 978-1-4503-4354-1.
\newblock \doi{10.1145/3020078.3021738}.
\newblock URL \url{http://doi.acm.org/10.1145/3020078.3021738}.

\bibitem[Banner et~al.(2018)Banner, Hubara, Hoffer, and
  Soudry]{DBLP:journals/corr/abs-1805-11046}
Ron Banner, Itay Hubara, Elad Hoffer, and Daniel Soudry.
\newblock Scalable methods for 8-bit training of neural networks.
\newblock \emph{CoRR}, abs/1805.11046, 2018.
\newblock URL \url{http://arxiv.org/abs/1805.11046}.

\bibitem[Bengio et~al.(2013)Bengio, L{\'{e}}onard, and
  Courville]{bengio2013estimating}
Yoshua Bengio, Nicholas L{\'{e}}onard, and Aaron~C. Courville.
\newblock Estimating or propagating gradients through stochastic neurons for
  conditional computation.
\newblock \emph{CoRR}, abs/1308.3432, 2013.
\newblock URL \url{http://arxiv.org/abs/1308.3432}.

\bibitem[Cai et~al.(2017)Cai, He, Sun, and Vasconcelos]{cai2017deep}
Zhaowei Cai, Xiaodong He, Jian Sun, and Nuno Vasconcelos.
\newblock Deep learning with low precision by half-wave gaussian quantization.
\newblock In \emph{IEEE Computer Society Conference on Computer Vision and
  Pattern Recognition}, 2017.

\bibitem[Chen et~al.(2018)Chen, Papandreou, Kokkinos, Murphy, and
  Yuille]{DBLP:journals/corr/ChenPK0Y16}
Liang{-}Chieh Chen, George Papandreou, Iasonas Kokkinos, Kevin Murphy, and
  Alan~L. Yuille.
\newblock Deeplab: Semantic image segmentation with deep convolutional nets,
  atrous convolution, and fully connected crfs.
\newblock \emph{IEEE Transactions on Pattern Analysis and Machine
  Intelligence}, 40\penalty0 (4):\penalty0 834--848, April 2018.
\newblock ISSN 0162-8828.
\newblock \doi{10.1109/TPAMI.2017.2699184}.

\bibitem[Chen et~al.(2016)Chen, Emer, and Sze]{7551407}
Yu-Hsin Chen, Joel Emer, and Vivienne Sze.
\newblock Eyeriss: A spatial architecture for energy-efficient dataflow for
  convolutional neural networks.
\newblock In \emph{2016 ACM/IEEE 43rd Annual International Symposium on
  Computer Architecture (ISCA)}, volume~44, pp.\  367--379, June 2016.
\newblock \doi{10.1109/ISCA.2016.40}.

\bibitem[Choi et~al.(2018)Choi, Wang, Venkataramani, Chuang, Srinivasan, and
  Gopalakrishnan]{choi2018pact}
Jungwook Choi, Zhuo Wang, Swagath Venkataramani, Pierce~I{-}Jen Chuang,
  Vijayalakshmi Srinivasan, and Kailash Gopalakrishnan.
\newblock {PACT:} parameterized clipping activation for quantized neural
  networks.
\newblock \emph{CoRR}, abs/1805.06085, 2018.
\newblock URL \url{http://arxiv.org/abs/1805.06085}.

\bibitem[Dong et~al.(2017)Dong, Ni, Li, Chen, Zhu, and Su]{dong2017learning}
Yinpeng Dong, Renkun Ni, Jianguo Li, Yurong Chen, Jun Zhu, and Hang Su.
\newblock Learning accurate low-bit deep neural networks with stochastic
  quantization.
\newblock In \emph{British Machine Vision Conference (BMVC'17)}, 2017.

\bibitem[Ghaffari \& Sharifian(2016)Ghaffari and Sharifian]{7869873}
Sina Ghaffari and Saeed Sharifian.
\newblock Fpga-based convolutional neural network accelerator design using high
  level synthesize.
\newblock In \emph{2016 2nd International Conference of Signal Processing and
  Intelligent Systems (ICSPIS)}, pp.\  1--6, Dec 2016.
\newblock \doi{10.1109/ICSPIS.2016.7869873}.

\bibitem[Gray(1990)]{gray1990quantization}
Robert~M Gray.
\newblock Quantization noise spectra.
\newblock \emph{IEEE Transactions on information theory}, 36\penalty0
  (6):\penalty0 1220--1244, 1990.

\bibitem[Gupta et~al.(2015)Gupta, Agrawal, Gopalakrishnan, and
  Narayanan]{gupta2015deep}
Suyog Gupta, Ankur Agrawal, Kailash Gopalakrishnan, and Pritish Narayanan.
\newblock Deep learning with limited numerical precision.
\newblock In \emph{Proceedings of the 32nd International Conference on Machine
  Learning (ICML-15)}, pp.\  1737--1746, 2015.

\bibitem[He et~al.(2016)He, Zhang, Ren, and Sun]{He_2016_CVPR}
Kaiming He, Xiangyu Zhang, Shaoqing Ren, and Jian Sun.
\newblock Deep residual learning for image recognition.
\newblock In \emph{The IEEE Conference on Computer Vision and Pattern
  Recognition (CVPR)}, June 2016.

\bibitem[Hinton et~al.(2012)Hinton, Deng, Yu, Dahl, Mohamed, Jaitly, Senior,
  Vanhoucke, Nguyen, Sainath, and Kingsbury]{6296526}
Geoffrey Hinton, Li~Deng, Dong Yu, George~E Dahl, Abdel-rahman Mohamed, Navdeep
  Jaitly, Andrew Senior, Vincent Vanhoucke, Patrick Nguyen, Tara~N Sainath, and
  Brian Kingsbury.
\newblock Deep neural networks for acoustic modeling in speech recognition: The
  shared views of four research groups.
\newblock \emph{IEEE Signal Processing Magazine}, 29\penalty0 (6):\penalty0
  82--97, Nov 2012.
\newblock ISSN 1053-5888.
\newblock \doi{10.1109/MSP.2012.2205597}.

\bibitem[Hinton et~al.(2015)Hinton, Vinyals, and Dean]{hinton2015distilling}
Geoffrey Hinton, Oriol Vinyals, and Jeffrey Dean.
\newblock Distilling the knowledge in a neural network.
\newblock 2015.
\newblock URL \url{http://arxiv.org/abs/1503.02531}.

\bibitem[Hubara et~al.(2016)Hubara, Courbariaux, Soudry, El-Yaniv, and
  Bengio]{hubara2016binarized}
Itay Hubara, Matthieu Courbariaux, Daniel Soudry, Ran El-Yaniv, and Yoshua
  Bengio.
\newblock Binarized neural networks.
\newblock In \emph{Advances in neural information processing systems}, pp.\
  4107--4115, 2016.

\bibitem[Hubara et~al.(2018)Hubara, Courbariaux, Soudry, El-Yaniv, and
  Bengio]{hubara2016quantized}
Itay Hubara, Matthieu Courbariaux, Daniel Soudry, Ran El-Yaniv, and Yoshua
  Bengio.
\newblock Quantized neural networks: Training neural networks with low
  precision weights and activations.
\newblock \emph{Journal of Machine Learning Research}, 18\penalty0
  (187):\penalty0 1--30, 2018.
\newblock URL \url{http://jmlr.org/papers/v18/16-456.html}.

\bibitem[Jacob et~al.(2018)Jacob, Kligys, Chen, Zhu, Tang, Howard, Adam, and
  Kalenichenko]{jacob2017quantization}
Benoit Jacob, Skirmantas Kligys, Bo~Chen, Menglong Zhu, Matthew Tang, Andrew
  Howard, Hartwig Adam, and Dmitry Kalenichenko.
\newblock Quantization and training of neural networks for efficient
  integer-arithmetic-only inference.
\newblock In \emph{The IEEE Conference on Computer Vision and Pattern
  Recognition (CVPR)}, June 2018.

\bibitem[Jouppi et~al.(2017)Jouppi, Young, Patil, Patterson, Agrawal, Bajwa,
  Bates, Bhatia, Boden, Borchers, Boyle, Cantin, Chao, Clark, Coriell, Daley,
  Dau, Dean, Gelb, Ghaemmaghami, Gottipati, Gulland, Hagmann, Ho, Hogberg, Hu,
  Hundt, Hurt, Ibarz, Jaffey, Jaworski, Kaplan, Khaitan, Koch, Kumar, Lacy,
  Laudon, Law, Le, Leary, Liu, Lucke, Lundin, MacKean, Maggiore, Mahony,
  Miller, Nagarajan, Narayanaswami, Ni, Nix, Norrie, Omernick, Penukonda,
  Phelps, Ross, Salek, Samadiani, Severn, Sizikov, Snelham, Souter, Steinberg,
  Swing, Tan, Thorson, Tian, Toma, Tuttle, Vasudevan, Walter, Wang, Wilcox, and
  Yoon]{DBLP:journals/corr/JouppiYPPABBBBB17}
Norman~P. Jouppi, Cliff Young, Nishant Patil, David~A. Patterson, Gaurav
  Agrawal, Raminder Bajwa, Sarah Bates, Suresh Bhatia, Nan Boden, Al~Borchers,
  Rick Boyle, Pierre{-}luc Cantin, Clifford Chao, Chris Clark, Jeremy Coriell,
  Mike Daley, Matt Dau, Jeffrey Dean, Ben Gelb, Tara~Vazir Ghaemmaghami,
  Rajendra Gottipati, William Gulland, Robert Hagmann, Richard~C. Ho, Doug
  Hogberg, John Hu, Robert Hundt, Dan Hurt, Julian Ibarz, Aaron Jaffey, Alek
  Jaworski, Alexander Kaplan, Harshit Khaitan, Andy Koch, Naveen Kumar, Steve
  Lacy, James Laudon, James Law, Diemthu Le, Chris Leary, Zhuyuan Liu, Kyle
  Lucke, Alan Lundin, Gordon MacKean, Adriana Maggiore, Maire Mahony, Kieran
  Miller, Rahul Nagarajan, Ravi Narayanaswami, Ray Ni, Kathy Nix, Thomas
  Norrie, Mark Omernick, Narayana Penukonda, Andy Phelps, Jonathan Ross, Amir
  Salek, Emad Samadiani, Chris Severn, Gregory Sizikov, Matthew Snelham, Jed
  Souter, Dan Steinberg, Andy Swing, Mercedes Tan, Gregory Thorson, Bo~Tian,
  Horia Toma, Erick Tuttle, Vijay Vasudevan, Richard Walter, Walter Wang, Eric
  Wilcox, and Doe~Hyun Yoon.
\newblock In-datacenter performance analysis of a tensor processing unit.
\newblock In \emph{2017 ACM/IEEE 44th Annual International Symposium on
  Computer Architecture (ISCA)}, pp.\  1--12, June 2017.
\newblock \doi{10.1145/3079856.3080246}.

\bibitem[Jung et~al.(2018)Jung, Son, Lee, Son, Kwak, Han, and
  Choi]{2018arXiv180805779J}
Sangil Jung, Changyong Son, Seohyung Lee, Jinwoo Son, Youngjun Kwak, Jae-Joon
  Han, and Changkyu Choi.
\newblock {Joint Training of Low-Precision Neural Network with Quantization
  Interval Parameters}.
\newblock \emph{ArXiv e-prints}, August 2018.

\bibitem[Khashabi et~al.(2014)Khashabi, Nowozin, Jancsary, and
  Fitzgibbon]{msrdemosaicing2014}
Daniel Khashabi, Sebastian Nowozin, Jeremy Jancsary, and Andrew~W. Fitzgibbon.
\newblock Joint demosaicing and denoising via learned nonparametric random
  fields.
\newblock \emph{IEEE Transactions on Image Processing}, 23\penalty0
  (12):\penalty0 4968--4981, 2014.
\newblock URL \url{http://dx.doi.org/10.1109/TIP.2014.2359774}.

\bibitem[Lai et~al.(2015)Lai, Xu, Liu, and Zhao]{Lai:2015:RCN:2886521.2886636}
Siwei Lai, Liheng Xu, Kang Liu, and Jun Zhao.
\newblock Recurrent convolutional neural networks for text classification.
\newblock In \emph{Proceedings of the Twenty-Ninth AAAI Conference on
  Artificial Intelligence}, AAAI'15, pp.\  2267--2273. AAAI Press, 2015.
\newblock ISBN 0-262-51129-0.
\newblock URL \url{http://dl.acm.org/citation.cfm?id=2886521.2886636}.

\bibitem[McKinstry et~al.(2018)McKinstry, Esser, Appuswamy, Bablani, Arthur,
  Yildiz, and Modha]{mckinstry2018disc}
Jeffrey~L. McKinstry, Steven~K. Esser, Rathinakumar Appuswamy, Deepika Bablani,
  John~V. Arthur, Izzet~B. Yildiz, and Dharmendra~S. Modha.
\newblock {Discovering Low-Precision Networks Close to Full-Precision Networks
  for Efficient Embedded Inference}.
\newblock \emph{ArXiv e-prints}, September 2018.

\bibitem[Mishra \& Marr(2018)Mishra and Marr]{mishra2017apprentice}
Asit Mishra and Debbie Marr.
\newblock Apprentice: Using knowledge distillation techniques to improve
  low-precision network accuracy.
\newblock \emph{International Conference on Learning Representations}, 2018.
\newblock URL \url{https://openreview.net/forum?id=B1ae1lZRb}.

\bibitem[Mishra et~al.(2018)Mishra, Nurvitadhi, Cook, and Marr]{mishra2017wrpn}
Asit Mishra, Eriko Nurvitadhi, Jeffrey~J Cook, and Debbie Marr.
\newblock Wrpn: Wide reduced-precision networks.
\newblock \emph{International Conference on Learning Representations}, 2018.
\newblock URL \url{https://openreview.net/forum?id=B1ZvaaeAZ}.

\bibitem[Polino et~al.(2018)Polino, Pascanu, and Alistarh]{polino2018model}
Antonio Polino, Razvan Pascanu, and Dan Alistarh.
\newblock Model compression via distillation and quantization.
\newblock \emph{International Conference on Learning Representations}, 2018.
\newblock URL \url{https://openreview.net/forum?id=S1XolQbRW}.

\bibitem[Rastegari et~al.(2016)Rastegari, Ordonez, Redmon, and
  Farhadi]{rastegari2016xnor}
Mohammad Rastegari, Vicente Ordonez, Joseph Redmon, and Ali Farhadi.
\newblock Xnor-net: Imagenet classification using binary convolutional neural
  networks.
\newblock In \emph{European Conference on Computer Vision}, pp.\  525--542.
  Springer, 2016.

\bibitem[Schwartz et~al.(2018)Schwartz, Giryes, and
  Bronstein]{schwartz2018deepisp}
Eli Schwartz, Raja Giryes, and Alex~M Bronstein.
\newblock Deepisp: Learning end-to-end image processing pipeline.
\newblock \emph{arXiv preprint arXiv:1801.06724}, 2018.

\bibitem[Sripad \& Snyder(1977)Sripad and Snyder]{sripad1977necessary}
Anekal Sripad and Donald Snyder.
\newblock A necessary and sufficient condition for quantization errors to be
  uniform and white.
\newblock \emph{IEEE Transactions on Acoustics, Speech, and Signal Processing},
  25\penalty0 (5):\penalty0 442--448, 1977.

\bibitem[Umuroglu et~al.(2017)Umuroglu, Fraser, Gambardella, Blott, Leong,
  Jahre, and Vissers]{DBLP:journals/corr/UmurogluFGBLJV16}
Yaman Umuroglu, Nicholas~J. Fraser, Giulio Gambardella, Michaela Blott, Philip
  Leong, Magnus Jahre, and Kees Vissers.
\newblock Finn: A framework for fast, scalable binarized neural network
  inference.
\newblock In \emph{Proceedings of the 2017 ACM/SIGDA International Symposium on
  Field-Programmable Gate Arrays}, FPGA '17, pp.\  65--74, New York, NY, USA,
  2017. ACM.
\newblock ISBN 978-1-4503-4354-1.
\newblock \doi{10.1145/3020078.3021744}.
\newblock URL \url{http://doi.acm.org/10.1145/3020078.3021744}.

\bibitem[Wang et~al.(2016{\natexlab{a}})Wang, An, and
  Xu]{DBLP:journals/corr/WangAX16}
Dong Wang, Jianjing An, and Ke~Xu.
\newblock Pipecnn: An opencl-based {FPGA} accelerator for large-scale
  convolution neuron networks.
\newblock \emph{CoRR}, abs/1611.02450, 2016{\natexlab{a}}.
\newblock URL \url{http://arxiv.org/abs/1611.02450}.

\bibitem[Wang et~al.(2016{\natexlab{b}})Wang, Qiao, Liu, Shan, Zhou, Luo, and
  Yang]{7848692}
Zhengrong Wang, Fei Qiao, Zhen Liu, Yuxiang Shan, Xunyi Zhou, Li~Luo, and
  Huazhong Yang.
\newblock Optimizing convolutional neural network on fpga under heterogeneous
  computing framework with opencl.
\newblock In \emph{2016 IEEE Region 10 Conference (TENCON)}, pp.\  3433--3438,
  Nov 2016{\natexlab{b}}.
\newblock \doi{10.1109/TENCON.2016.7848692}.

\bibitem[Xu et~al.(2018)Xu, Wang, Zhou, Lin, and Xiong]{AAAI1816479}
Yuhui Xu, Yongzhuang Wang, Aojun Zhou, Weiyao Lin, and Hongkai Xiong.
\newblock Deep neural network compression with single and multiple level
  quantization, 2018.
\newblock URL
  \url{https://www.aaai.org/ocs/index.php/AAAI/AAAI18/paper/view/16479}.

\bibitem[Zhang et~al.(2018)Zhang, Yang, Ye, and Hua]{ZhangYangYeECCV2018}
Dongqing Zhang, Jiaolong Yang, Dongqiangzi Ye, and Gang Hua.
\newblock Lq-nets: Learned quantization for highly accurate and compact deep
  neural networks.
\newblock In \emph{European Conference on Computer Vision (ECCV)}, 2018.

\bibitem[Zhao et~al.(2017)Zhao, Song, Zhang, Xing, Lin, Srivastava, Gupta, and
  Zhang]{Zhao:2017:ABC:3020078.3021741}
Ritchie Zhao, Weinan Song, Wentao Zhang, Tianwei Xing, Jeng-Hau Lin, Mani
  Srivastava, Rajesh Gupta, and Zhiru Zhang.
\newblock Accelerating binarized convolutional neural networks with
  software-programmable fpgas.
\newblock In \emph{Proceedings of the 2017 ACM/SIGDA International Symposium on
  Field-Programmable Gate Arrays}, FPGA '17, pp.\  15--24, New York, NY, USA,
  2017. ACM.
\newblock ISBN 978-1-4503-4354-1.
\newblock \doi{10.1145/3020078.3021741}.
\newblock URL \url{http://doi.acm.org/10.1145/3020078.3021741}.

\bibitem[Zhou et~al.(2017)Zhou, Yao, Guo, Xu, and Chen]{zhou2017incremental}
Aojun Zhou, Anbang Yao, Yiwen Guo, Lin Xu, and Yurong Chen.
\newblock Incremental network quantization: Towards lossless cnns with
  low-precision weights.
\newblock In \emph{International Conference on Learning
  Representations,ICLR2017}, 2017.

\bibitem[Zhou et~al.(2016)Zhou, Wu, Ni, Zhou, Wen, and Zou]{zhou2016dorefa}
Shuchang Zhou, Yuxin Wu, Zekun Ni, Xinyu Zhou, He~Wen, and Yuheng Zou.
\newblock Dorefa-net: Training low bitwidth convolutional neural networks with
  low bitwidth gradients.
\newblock \emph{arXiv preprint arXiv:1606.06160}, 2016.

\bibitem[Zhu et~al.(2016)Zhu, Han, Mao, and Dally]{zhu2016trained}
Chenzhuo Zhu, Song Han, Huizi Mao, and William~J Dally.
\newblock Trained ternary quantization.
\newblock \emph{International Conference on Learning Representations (ICLR)},
  2016.

\end{thebibliography}
	\bibliographystyle{iclr2019_conference}
	
  \newpage
  \appendix

  \renewcommand\thefigure{\thesection.\arabic{figure}} 
  \renewcommand\thetable{\thesection.\arabic{table}}

  \section{Quantization error distribution}
  \label{sec:quant_err_dist}
    \setcounter{figure}{0}  
  \setcounter{table}{0}  
  
 	\begin{figure*}[t]
		\vskip 0in
		\begin{center}
			\centerline{\includegraphics[width=\linewidth]{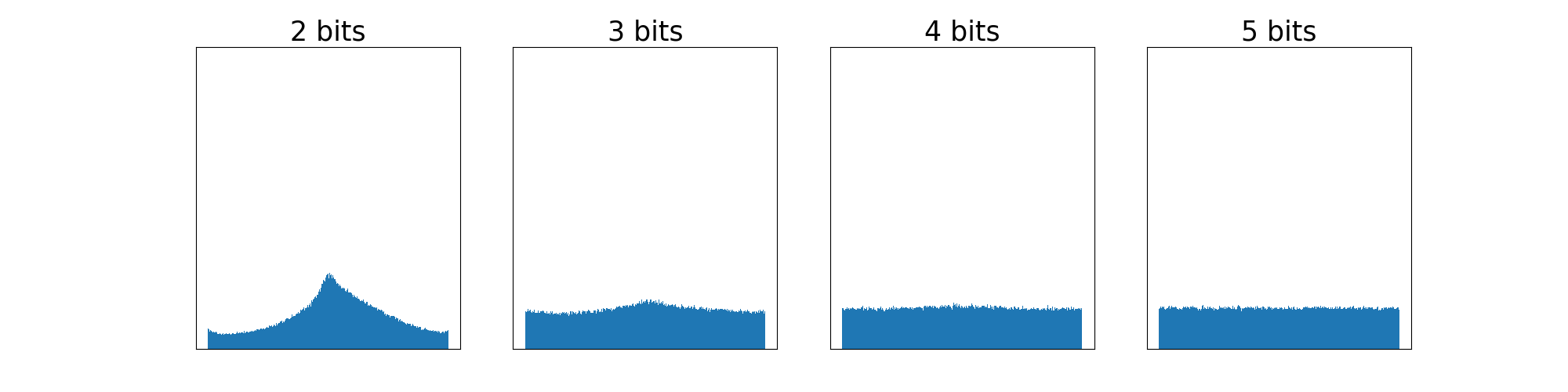}}
			\vspace{-0.3cm}
			\caption{Weight quantization error histogram for a range of bitwidths }
			\label{quant_error}
		\end{center}
		\vskip -0.2in
	\end{figure*}
 
 The general statement is that for large number of bins, the distribution of quantization error is independent on the quantized value, and thus distributed uniformly. However, this is true only in limit of high number of bins, which is not exactly the case of neural network quantization. However, empirically the distribution of noise is almost uniform for 4 and 5 bits and only starts to deviate deviating from the uniform model (Figure \ref{quant_error}) for 3 bits, which corresponds to only 8 bins. 
  
  \section{Clamping parameter convergence}
    \label{act_clamp_app}
\setcounter{figure}{0}  
  \setcounter{table}{0}

    	\begin{figure*}[t]
		\vskip 0in
		\begin{center}
			\centerline{\includegraphics[width=0.5\linewidth]{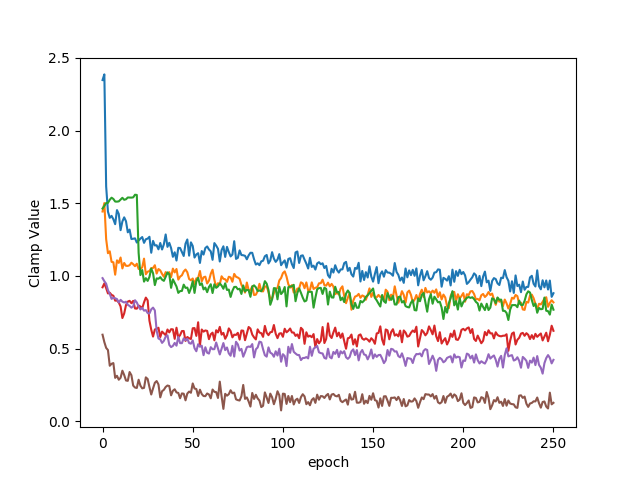}}
			\vspace{-0.3cm}
			\caption{
				 Activation clamp values during ResNet-18 training on CIFAR10 dataset  }
			\label{activation_clamp}
		\end{center}
		\vskip -0.2in
	\end{figure*}
    
    	Figure \ref{activation_clamp} depicts the evolution of the activation clamp values throughout the epochs. In this experiment $\alpha$ was set to 5. It can be seen that activation clamp values converge to values smaller than the initialization. This shows that the layer prefers to shrink the dynamic range of the activations, which can be interpreted as a form of regularization similar in its purpose to weight decay on weights.
   
   \section{Experiments on CIFAR-10}
  \label{cifar_app}
  \setcounter{figure}{0}  
  \setcounter{table}{0}  
	As an additional experiment,     we test \uniqname with ResNet-18 on CIFAR-10 for various quantization levels of the weights and activations. Table~\ref{tab_cifar_dif_bitwidth} reports the results. Notice that for the case of 3-bit weights  activations we get the same accuracy and for the 2-bit case only a small degradation. Moreover, 
observe that when we quantize only the weights or activations, we get a nice regularization effect that  improves the achieved accuracy.

	\begin{table}[th]
		\centering
			\caption{\uniqname accuracy (\% top-1)  on CIFAR-10 for range of bitwidths.}
			\label{tab_cifar_dif_bitwidth}
			\begin{center}
				\begin{small}
					\begin{tabular}{cc|cccc}
						\toprule
						&    & \multicolumn{3}{c}{\bf{Activation bits}}                         \\
						&    & 1                                        & 2     & 3     & 32    \\
						\midrule
						\multirow{3}{*}{\rotatebox{90}{\makecell{\bf{Weight}                     \\\bf{bits}}}}
						& 2  & 89.5                                     & 92.53 & 92.69 & 92.71 \\
						& 3  & 91.32                                        &92.74     & 93.01 & 93.26 \\
						& 32 & 91.87                                        & 93.04     &  93.15    & 93.02 \\
						\bottomrule
					\end{tabular}
				\end{small}
			\end{center}
		\vspace{-0.45cm}
	\end{table}
    
  \section{Background for Neural Networks on custom hardware}
  \label{hardware_app}
  \setcounter{figure}{0}  
  \setcounter{table}{0}  
	When implementing systems involving arbitrary precision,
	FPGAs and ASICs are a natural selection as target device
	due to their customizable nature. It was already shown that there is a lot of redundancy when using floating point representation in Neural Network(NN). Therefore, custom low-precision representation can be used with little impact to the accuracy. Due to the steadily increasing on-chip memory
	size (tens of megabytes) and the integration of high bandwidth memory (hundreds of megabytes), it is feasible to fit
	all the parameters inside an ASIC or FPGA, when using low bitwidth. Besides the obvious advantage of reducing the
	latency, this approach has several advantages: power consumption reduction and smaller resource utilization, which
	in addition to DSP blocks and LUTs, also includes routing
	resource. The motivation of quantizing the activations is
	similar to that of the parameters. Although activations are
	not stored during inference, their quantization can lead to major saving
	in routing resources which in turn can increase the maximal
	operational frequency of the fabric, resulting in increased
	throughput.
		In recent years, FPGAs has become more popular as an inference accelerator. And while ASICs \cite{7551407,DBLP:journals/corr/JouppiYPPABBBBB17} usually offers more throughput with lower energy consumption, they don't enjoy the advantage of reconfigurability as FPGAs. This is important since neural network algorithm evolve with time, so should their hardware implementation. Since the implementation of neural network involves complex scheduling and data movement, FPGA-based inference accelerators has been described as heterogeneous system using OpenCL \cite{DBLP:journals/corr/WangAX16,7848692,DBLP:journals/corr/AydonatOCLC17} or as standalone accelerator using HLS compilers \cite{7869873,DBLP:journals/corr/UmurogluFGBLJV16,Zhao:2017:ABC:3020078.3021741}.
\end{document}